\documentclass{article}

    \usepackage[nonatbib, preprint]{neurips_2019}

\usepackage[utf8]{inputenc} 
\usepackage[T1]{fontenc}    
\usepackage{hyperref}       
\usepackage{url}            
\usepackage{booktabs}       
\usepackage{amsfonts}       
\usepackage{nicefrac}       
\usepackage{microtype}      
\usepackage{graphicx}
\usepackage{multirow}

\title{Transformer-based normative modelling for anomaly detection of early schizophrenia}

\author{%
Pedro F Da Costa$^{1,2}$ \quad Jessica Dafflon$^{3,4}$ \quad Sergio Leonardo Mendes$^5$ \\ 
\textbf{João Ricardo Sato}$^5$ \quad \textbf{M. Jorge Cardoso}$^6$ \quad  \textbf{Robert Leech}$^1$ \\
\textbf{Emily JH Jones}$^2$ \quad \textbf{Walter H. L. Pinaya }$^6$ \\
\\
$^1$Institute of Psychiatry, Psychology $\&$ Neuroscience, King's College London, UK \\ 
$^2$Centre for Brain and Cognitive Development, Birkbeck College, London, UK \\
$^3$Data Science and Sharing Team, National Institute of Mental Health, Bethesda, MD, USA \\
$^4$Machine Learning Team, National Institute of Mental Health, Bethesda, MD, USA \\
$^5$Center of Mathematics, Computing, and Cognition, Universidade Federal do ABC, Brazil \\
$^6$School of Biomedical Engineering $\&$ Imaging Sciences, King’s College London, UK \\
}

\begin{document}

\maketitle

\begin{abstract}
Despite the impact of psychiatric disorders on clinical health, early-stage diagnosis remains a challenge. Machine learning studies have shown that classifiers tend to be overly narrow in the diagnosis prediction task. The overlap between conditions leads to high heterogeneity among participants that is not adequately captured by classification models. To address this issue, normative approaches have surged as an alternative method. By using a generative model to learn the distribution of healthy brain data patterns, we can identify the presence of pathologies as deviations or outliers from the distribution learned by the model. In particular, deep generative models showed great results as normative models to identify neurological lesions in the brain. However, unlike most neurological lesions, psychiatric disorders present subtle changes widespread in several brain regions, making these alterations challenging to identify. In this work, we evaluate the performance of transformer-based normative models to detect subtle brain changes expressed in adolescents and young adults. We trained our model on 3D MRI scans of neurotypical individuals (N=1,765). Then, we obtained the likelihood of neurotypical controls and psychiatric patients with early-stage schizophrenia from an independent dataset (N=93) from the Human Connectome Project. Using the predicted likelihood of the scans as a proxy for a normative score, we obtained an AUROC of 0.82 when assessing the difference between controls and individuals with early-stage schizophrenia. Our approach surpassed recent normative methods based on brain age and Gaussian Process, showing the promising use of deep generative models to help in individualised analyses.
\end{abstract}

\section{Introduction}

Schizophrenia is a chronic mental health disorder that causes a range of heterogenic psychological symptoms and significantly impairs the quality of life of millions of people worldwide. The current gold-standard diagnosis for schizophrenia, described in the Diagnostics and Statistical Manual of Mental Disorders (DSM-V) \cite{edition2013diagnostic}) relies on a professional inquiring the patient and evaluating the presence of at least three of the five main symptoms (i.e., delusions, hallucinations, disorganised or incoherent speaking, disorganised or unusual movements and negative symptoms). Issues of inter-rater reliability and ambiguous criteria descriptions \cite{welch2013dsm} are factors that are pushing the field of psychiatry to search for more objective and operationalisable biomarkers of psychiatric conditions.

It has been shown that schizophrenia is associated with subtle brain abnormalities that can be detected via structural Magnetic Resonance Imaging (sMRI) data \cite{Shenton2001}. This has led researchers to try to build reliable predictors of schizophrenia by employing machine learning algorithms. The most popular approaches utilised supervised learning on structural data to build classifiers \cite{squarcina2017,Leonard1999}. Despite presenting modest to good accuracies on the testing sets, most algorithms fail to generalize to the early stages of the condition and to cross-site validation \cite{Pinaya2016,Vieira2019}. These failures can be partly attributed to training on small and limited datasets that fail to capture the full distribution of patients and the population in general. 

To address these limitations, there has been an effort to move to unsupervised learning techniques that focus on building normative models of the healthy brain and try to capture variations from normality as predictors of psychiatric conditions \cite{Marquand2019}. The challenge of collecting large swathes of brain data becomes more manageable if only healthy participants' data are required. 
Recently, researchers have proposed unsupervised anomaly detection algorithms to identify brain pathologies, such as brain lesions, from sMRI \cite{baur2018deep,baur2021autoencoders,chen2020unsupervised}. These methods are based on autoencoders to learn a latent representation of healthy brain data. After training, these models assess new examples to detect pathologies based on their deviation from normality. In this context, the current state of the art is held by variational autoencoder (VAE) based methods \cite{baur2020scale}, which try to reconstruct a test image as the nearest sample on the learned normal manifold, using the reconstruction error to quantify the degree and spatial distribution of any anomaly \cite{pinaya2021unsupervised}. However, the success of this approach is limited by the fidelity of reconstructions from most VAE architectures \cite{dumoulin2016adversarially}, and by unwanted reconstructions of pathological features not present in training data, which suggests a failure of the model to internalize complex relationships between distant imaging features \cite{pinaya2021unsupervised}. To address these issues, a recent study achieved state-of-the-art performance in unsupervised brain anomaly detection using an architecture based on transformers \cite{pinaya2021unsupervised}. The robustness of transformers to map input data relationships, whose distances vary widely, makes them great candidates for neuroimaging tasks, especially anomaly detection \cite{graham2022transformer,pinaya2021unsupervised}.

In this study, we investigated if a normative model with an architecture based on transformers could be used to detect psychopathologies, such as schizophrenia, from brain 3d sMRI and if it could be further used to study the local variations associated with this condition. We found that by training a Vector Quantized VAE (VQVAE) with an autoregressive transformer, we could classify between controls and participants with early-stage schizophrenia with high accuracy.

\section{Methods}

\subsection{Normative model}

The main component of the algorithm is an autoregressive Transformer \cite{vaswani2017attention} that learns a mapping of probabilities of a given sequence of values, which is an approximation of the likelihood of the distribution. Although computationally demanding, Transformers are able to capture highly complex dependencies across large distances due to their attention mechanism that weighs the linear transformation of the input with itself. The computational cost of the attention mechanism scales quadratically with the sequence length, making it unfeasible to sequence the original highly-dimensional brain data. Our method encodes the 3-dimensional sMRI brain data into a smaller discrete latent space before being fed as input to the autoregressive Transformer. Using a VQVAE \cite{VanDenOord2017}, we reduce the dimensionality from 192x224x192 to 24x28x24 voxels while maintaining the relevant structural information. This dimensionality-reduction step makes it feasible to train a Transformer to learn the probability distribution of the latent space. Both the VQVAE and the Transformer are trained separately using only healthy participants from the Human Connectome Project. Further details on the data used, model architecture and training settings are described in the supplementary material.

\subsection{Evaluation and analysis}
We estimate the probability of each index in the latent representation obtained by the VQVAE for each brain scan through the output of the trained transformer. The autoregressive transformer weighs the already evaluated latent indices to predict the conditional probability of the next index in the latent representation. This method will result in the model flagging indices that do not follow the sequencing observed from the trained data by assigning them low probability values. This follows from cases where there are changes to the normal structuring of the brain as the transformer was trained on a healthy cohort. By summing the log-likelihood of all the elements of the latent representation, we obtain a log-likelihood estimation per individual in the evaluation set. The evaluation set is composed by the Human Connectome Project for Early Psychosis (HCP-EP) \footnote[1]{https://www.humanconnectome.org/study/human-connectome-project-for-early-psychosis/document/hcp-ep} cohort, where both neurotypical participants (n=46), and participants with early-stage schizophrenia (n=47) are present. We hypothesise that images acquired from participants with schizophrenia will have a lower log-likelihood than controls due to the subtle changes in their brain structures being captured by the transformer as unlikely. We measure this through a correlation analysis between the participants diagnosis and its log-likelihood. We further study how efficient this method is at identifying individuals with early-stage schizophrenia by measuring the Area Under the ROC curve (AUROC) of the log-likelihood estimation with the diagnosis as the target variable. As baseline, we evaluated three normative models that have previously been considered for psychiatry data: Gaussian process regression \cite{Marquand2016}, Bayesian linear regression \cite{HUERTAS2017134} and a state-of-the-art brain age prediction model \cite{Peng2021}. For further details on these baselines, see the supplementary materials.
Finally, we studied how likelihood measures vary locally for participants with schizophrenia by mapping the likelihood estimations of the latent representations to different regions of the brain space, using the Desikan-Killiany cortical atlas (aparc) and the automatic segmentation volume (aseg) \cite{Desikan2006}.

\section{Results}
As shown in Table \ref{tab:performance}, our model outperforms all baselines at identifying participants with schizophrenia showing a high AUROC of 0.828. We also obtained a Pearson's correlation coefficient of 0.568 (p-value < 0.001) when analyzing the correlation between the brain log-likelihood and the categorical label of diagnosis. At a regional level of analysis, no single region's likelihood presented a correlation to the diagnosis higher than the global likelihood estimation, but the prefrontal cortex, the temporal cortex and the ventricles presented the highest effect size measured  through Cohen's D.

\begin{table}
\caption{Performance of the different normative methods at identifying participants with schizophrenia as outliers. All methods presented a significant difference between the distribution of controls and individuals with schizophrenia (i.e., p-value < 0.05) for their metrics metric of normality.  }\label{tab:performance}
\begin{center}
\begin{tabular}{l c c c}
\hline
Method                      & AUROC $\uparrow$      & Correlation $\uparrow$    & P-value \\ 
\hline
GPR \cite{Marquand2016}       & 0.636                  & 0.215                     & 0.038    \\ 
BLR \cite{HUERTAS2017134}       & 0.678                  & 0.301                     & 0.003    \\

Brain Age \cite{Peng2021}             & 0.732                  & 0.391                     & <0.001 \\ 
VQVAE + Transformer [\textbf{Ours}]  & \textbf{0.828}                  & \textbf{0.568}                     & <0.001 \\
\hline
\end{tabular}
\end{center}
\end{table}

\begin{figure}
\centering
\includegraphics[width=\textwidth]{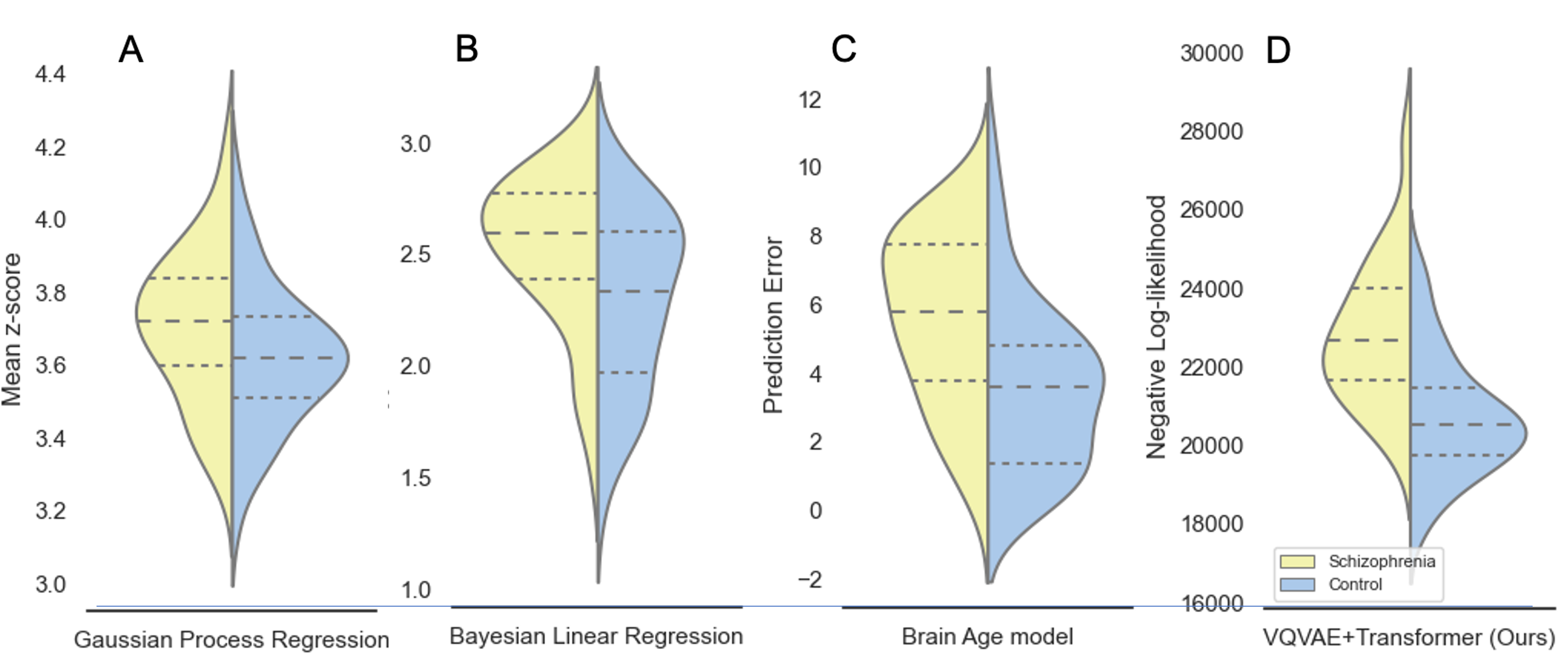}
\caption{\textbf{Violin plots of proxies used for anomaly detection}. For each distribution, the dashed lines identifies the median and the 25th and 75th quartiles. A. presents the distribution between cohorts for the mean z-scores obtained using uni-variate Gaussian process regressions. B. presents the distributions of the mean z-scores using Bayesian linear regression. C. presents the distributions obtained using the mean absolute error of a brain age model and D. presents our model's difference in distribution between control participants and individuals with early-stage schizophrenia using the estimated negative log likelihood per individual as a metric for detecting schizophrenia.}
\label{fig:fig1}
\end{figure}

\section{Discussion} \label{discussion}

In our study, we applied VQVAE and autoregressive transformers to create a normative method to predict how likely it is for a sample to belong to the healthy population. When applying the method to subjects from the HCP-EP dataset, we showed that our method was able to distinguish between healthy controls and participants with early-stage schizophrenia with a AUCROC=0.828. By using data from participants within 3 years of the onset of psychotic symptoms, we are less exposed to confounds such as prolonged medication or chronicity. Our normative score was more robust compared to other baseline methods, such as the z-score from the fitted Gaussian process regression, the z-score from the fitted Bayesian linear regression, or the prediction error of the state-of-the-art in brain age prediction (the Simple Fully Convolution Network (SFCN) \cite{Peng2021} model). Please refer to the Supplementary material for more information about the different models.  

The Gaussian process and Bayesian linear regression underperform the other two approaches in identifying schizophrenic participants as outliers. Fitting as many models as there are local regions makes these two normative approaches incapable of retrieving information among brain regions when building uncertainty estimates. As schizophrenia is known to be a condition that affects several areas simultaneously and the way that these areas communicate with each other \cite{tsuang1990, Lv2021}, it is expected that these models underperform. This is not the case in both the brain age approach and our VQVAE+Transformer model, where the voxel-level data is used to train both models in a multivariate method, where deep neural networks' high non-linearity can extract high-level information from different regions at the same time in a fully data-driven manner. One reason that can justify our model outperforming the brain age approach is that learning to map the age of individuals through MRI brain data results in using the reconstruction error as a proxy for identifying outliers which is an indirect task. Our model learns to explicitly predict the likelihood distribution of brain data from healthy participants. Another advantage of our model against brain age modeling is that it outputs a regional estimation of the likelihood that can be used to analyse the local structures for biomarkers of a given anomaly. There is no direct local representation for the brain age model as it predicts just a single value (i.e., the age of the participant).

One important limitation of our method is that it does not account for the demographic information of the participants, such as age and sex when building the normative model. One avenue that we leave for future work is to condition the transformer estimation based on context to have a likelihood estimation that is demographic dependent. Another limitation of our approach is the robustness of the model on data that do not follow a similar distribution of the training data \cite{Molina2017}. In this scenario, the model fails to identify healthy brains, marking all data with low likelihood. This shift in distribution can happen when using different scanner's acquisition settings or different field strengths. In this work, an external dataset is used (i.e., the HCP-EP), but it is under the umbrella of the Human Connectome Project, as is the training data. One solution that was implemented here and can be further explored is the data augmentation of the training data to simulate the output of the different scanner's settings, increasing the model's robustness to distribution shifts.

Regarding the regional analysis, it was observed that no isolated region is driving the correct prediction of participants with schizophrenia. Therefore, taking the whole brain likelihood estimation benefits the prediction power. This assessment follows closely with what is known from the literature, as schizophrenia is a condition with heterogeneous brain structure profiles \cite{tsuang1990,Lv2021}. The regions with the largest effect sizes follow what is known from the literature about brain changes related to schizophrenia. The prefrontal cortex has been associated with the mechanisms that underlie the onset of the psychiatric conditions \cite{ZHOU2005,SHEPHERD2012,Rimol2010} and the temporal cortex has been associated with volume reduction \cite{SHEPHERD2012} and positive symptoms \cite{Walton2017} in participants with schizophrenia.

Due to the fast pace of the development of deep generative models in other fields like computer vision, we believe these models are a promising tool to help build stronger biomarkers for psychiatric research.

\subsubsection*{Acknowledgments}
PFDC is supported by the European Union’s HORIZON 2020 Research and Innovation Programme under the Marie Sklodowska-Curie Grant Agreement No 814302. JD is supported by the Intramural Research Program of the NIMH (ZIC-MH002960 and ZIC-MH002968). WHLP and MJC are supported by Wellcome Innovations [WT213038/Z/18/Z]. JRS and SLM are supported by São Paulo Research Foundation (FAPESP) Grants \#2018/21934-5, \#2018/04654-9 and \#2022/07782-3.

\bibliographystyle{splncs04}
\bibliography{mybibliography}

\begin{thebibliography}{10}
\providecommand{\url}[1]{\texttt{#1}}
\providecommand{\urlprefix}{URL }
\providecommand{\doi}[1]{https://doi.org/#1}

\bibitem{Pinaya2022}
{Unsupervised brain imaging 3D anomaly detection and segmentation with
  transformers}. Medical Image Analysis  \textbf{79},  102475 (jul 2022).
  \doi{10.1016/J.MEDIA.2022.102475}

\bibitem{ants}
Avants, B., Epstein, C., Grossman, M., Gee, J.: Symmetric diffeomorphic image
  registration with cross-correlation: Evaluating automated labeling of elderly
  and neurodegenerative brain. Medical Image Analysis  \textbf{12}(1),  26--41
  (2008). \doi{10.1016/j.media.2007.06.004},
  \url{http://www.sciencedirect.com/science/article/pii/S1361841507000606}

\bibitem{baur2021autoencoders}
Baur, C., Denner, S., Wiestler, B., Navab, N., Albarqouni, S.: Autoencoders for
  unsupervised anomaly segmentation in brain mr images: a comparative study.
  Medical Image Analysis  \textbf{69},  101952 (2021)

\bibitem{baur2018deep}
Baur, C., Wiestler, B., Albarqouni, S., Navab, N.: Deep autoencoding models for
  unsupervised anomaly segmentation in brain mr images. In: International
  MICCAI brainlesion workshop. pp. 161--169. Springer (2018)

\bibitem{baur2020scale}
Baur, C., Wiestler, B., Albarqouni, S., Navab, N.: Scale-space autoencoders for
  unsupervised anomaly segmentation in brain mri. In: International Conference
  on Medical Image Computing and Computer-Assisted Intervention. pp. 552--561.
  Springer (2020)

\bibitem{chen2020unsupervised}
Chen, X., You, S., Tezcan, K.C., Konukoglu, E.: Unsupervised lesion detection
  via image restoration with a normative prior. Medical image analysis
  \textbf{64},  101713 (2020)

\bibitem{Cole2019}
Cole, J.H., Marioni, R.E., Harris, S.E., Deary, I.J.: {Brain age and other
  bodily ‘ages': implications for neuropsychiatry}. Molecular Psychiatry
  \textbf{24}(2),  266--281 (2019). \doi{10.1038/s41380-018-0098-1},
  \url{https://doi.org/10.1038/s41380-018-0098-1}

\bibitem{Desikan2006}
Desikan, R.S., S{\'{e}}gonne, F., Fischl, B., Quinn, B.T., Dickerson, B.C.,
  Blacker, D., Buckner, R.L., Dale, A.M., Maguire, R.P., Hyman, B.T., Albert,
  M.S., Killiany, R.J.: {An automated labeling system for subdividing the human
  cerebral cortex on MRI scans into gyral based regions of interest.}
  NeuroImage  \textbf{31}(3),  968--980 (jul 2006).
  \doi{10.1016/j.neuroimage.2006.01.021}

\bibitem{dumoulin2016adversarially}
Dumoulin, V., Belghazi, I., Poole, B., Mastropietro, O., Lamb, A., Arjovsky,
  M., Courville, A.: Adversarially learned inference. arXiv preprint
  arXiv:1606.00704  (2016)

\bibitem{dunn1961multiple}
Dunn, O.J.: Multiple comparisons among means. Journal of the American
  statistical association  \textbf{56}(293),  52--64 (1961)

\bibitem{edition2013diagnostic}
Edition, F., et~al.: Diagnostic and statistical manual of mental disorders. Am
  Psychiatric Assoc  \textbf{21}(21),  591--643 (2013)

\bibitem{graham2022transformer}
Graham, M.S., Tudosiu, P.D., Wright, P., Pinaya, W.H.L., Jean-Marie, U., Mah,
  Y., Teo, J., J{\"a}ger, R.H., Werring, D., Nachev, P., et~al.:
  Transformer-based out-of-distribution detection for clinically safe
  segmentation. arXiv preprint arXiv:2205.10650  (2022)

\bibitem{HUERTAS2017134}
Huertas, I., Oldehinkel, M., {van Oort}, E.S., Garcia-Solis, D., Mir, P.,
  Beckmann, C.F., Marquand, A.F.: A bayesian spatial model for neuroimaging
  data based on biologically informed basis functions. NeuroImage
  \textbf{161},  134--148 (2017).
  \doi{https://doi.org/10.1016/j.neuroimage.2017.08.009},
  \url{https://www.sciencedirect.com/science/article/pii/S1053811917306560}

\bibitem{Leonard1999}
Leonard, C.M., Kuldau, J.M., Breier, J.I., Zuffante, P.A., Gautier, E.R.,
  Heron, D.C., Lavery, E.M., Packing, J., Williams, S.A., DeBose, C.A.:
  {Cumulative effect of anatomical risk factors for schizophrenia: an MRI
  study.} Biological psychiatry  \textbf{46}(3),  374--382 (aug 1999).
  \doi{10.1016/s0006-3223(99)00052-9}

\bibitem{Lv2021}
Lv, J., {Di Biase}, M., Cash, R.F.H., Cocchi, L., Cropley, V.L., Klauser, P.,
  Tian, Y., Bayer, J., Schmaal, L., Cetin-Karayumak, S., Rathi, Y., Pasternak,
  O., Bousman, C., Pantelis, C., Calamante, F., Zalesky, A.: {Individual
  deviations from normative models of brain structure in a large
  cross-sectional schizophrenia cohort}. Molecular Psychiatry  \textbf{26}(7),
  3512--3523 (2021). \doi{10.1038/s41380-020-00882-5},
  \url{https://doi.org/10.1038/s41380-020-00882-5}

\bibitem{Marquand2019}
Marquand, A.F., Kia, S.M., Zabihi, M., Wolfers, T., Buitelaar, J.K., Beckmann,
  C.F.: {Conceptualizing mental disorders as deviations from normative
  functioning}. Molecular Psychiatry  \textbf{24}(10),  1415--1424 (2019).
  \doi{10.1038/s41380-019-0441-1},
  \url{https://doi.org/10.1038/s41380-019-0441-1}

\bibitem{Marquand2016}
Marquand, A.F., Wolfers, T., Mennes, M., Buitelaar, J., Beckmann, C.F.: {Beyond
  Lumping and Splitting: A Review of Computational Approaches for Stratifying
  Psychiatric Disorders}. Biological Psychiatry: Cognitive Neuroscience and
  Neuroimaging  \textbf{1}(5),  433--447 (2016).
  \doi{https://doi.org/10.1016/j.bpsc.2016.04.002},
  \url{https://www.sciencedirect.com/science/article/pii/S2451902216300301}

\bibitem{Molina2017}
Molina, D., P{\'{e}}rez-Beteta, J., Mart{\'{i}}nez-Gonz{\'{a}}lez, A., Martino,
  J., Velasquez, C., Arana, E., P{\'{e}}rez-Garc{\'{i}}a, V.M.: {Lack of
  robustness of textural measures obtained from 3D brain tumor MRIs impose a
  need for standardization.} PloS one  \textbf{12}(6),  e0178843 (2017).
  \doi{10.1371/journal.pone.0178843}

\bibitem{Peng2021}
Peng, H., Gong, W., Beckmann, C.F., Vedaldi, A., Smith, S.M.: {Accurate brain
  age prediction with lightweight deep neural networks}. Medical Image Analysis
   \textbf{68},  101871 (2021).
  \doi{https://doi.org/10.1016/j.media.2020.101871},
  \url{https://www.sciencedirect.com/science/article/pii/S1361841520302358}

\bibitem{Pinaya2016}
Pinaya, W.H.L., Gadelha, A., Doyle, O.M., Noto, C., Zugman, A., Cordeiro, Q.,
  Jackowski, A.P., Bressan, R.A., Sato, J.R.: {Using deep belief network
  modelling to characterize differences in brain morphometry in schizophrenia}.
  Scientific Reports  \textbf{6}(1),  38897 (2016). \doi{10.1038/srep38897},
  \url{https://doi.org/10.1038/srep38897}

\bibitem{pinaya2021unsupervised}
Pinaya, W.H.L., Tudosiu, P.D., Gray, R., Rees, G., Nachev, P., Ourselin, S.,
  Cardoso, M.J.: Unsupervised brain anomaly detection and segmentation with
  transformers. arXiv preprint arXiv:2102.11650  (2021)

\bibitem{Rimol2010}
Rimol, L.M., Hartberg, C.B., Nesv{\aa}g, R., Fennema-Notestine, C., Hagler,
  D.J.J., Pung, C.J., Jennings, R.G., Haukvik, U.K., Lange, E., Nakstad, P.H.,
  Melle, I., Andreassen, O.A., Dale, A.M., Agartz, I.: {Cortical thickness and
  subcortical volumes in schizophrenia and bipolar disorder.} Biological
  psychiatry  \textbf{68}(1),  41--50 (jul 2010).
  \doi{10.1016/j.biopsych.2010.03.036}

\bibitem{Shenton2001}
Shenton, M.E., Dickey, C.C., Frumin, M., McCarley, R.W.: {A review of MRI
  findings in schizophrenia}. Schizophrenia Research  \textbf{49}(1-2),  1--52
  (apr 2001). \doi{10.1016/S0920-9964(01)00163-3}

\bibitem{SHEPHERD2012}
Shepherd, A.M., Laurens, K.R., Matheson, S.L., Carr, V.J., Green, M.J.:
  {Systematic meta-review and quality assessment of the structural brain
  alterations in schizophrenia}. Neuroscience and Biobehavioral Reviews
  \textbf{36}(4),  1342--1356 (2012).
  \doi{https://doi.org/10.1016/j.neubiorev.2011.12.015},
  \url{https://www.sciencedirect.com/science/article/pii/S0149763411002223}

\bibitem{somerville2018lifespan}
Somerville, L.H., Bookheimer, S.Y., Buckner, R.L., Burgess, G.C., Curtiss,
  S.W., Dapretto, M., Elam, J.S., Gaffrey, M.S., Harms, M.P., Hodge, C.,
  et~al.: The lifespan human connectome project in development: A large-scale
  study of brain connectivity development in 5--21 year olds. Neuroimage
  \textbf{183},  456--468 (2018)

\bibitem{squarcina2017}
Squarcina, L., Castellani, U., Bellani, M., Perlini, C., Lasalvia, A., Dusi,
  N., Bonetto, C., Cristofalo, D., Tosato, S., Rambaldelli, G., Alessandrini,
  F., Zoccatelli, G., Pozzi-Mucelli, R., Lamonaca, D., Ceccato, E., Pileggi,
  F., Mazzi, F., Santonastaso, P., Ruggeri, M., Brambilla, P.: {Classification
  of first-episode psychosis in a large cohort of patients using support vector
  machine and multiple kernel learning techniques}. NeuroImage  \textbf{145},
  238--245 (2017). \doi{https://doi.org/10.1016/j.neuroimage.2015.12.007},
  \url{https://www.sciencedirect.com/science/article/pii/S1053811915011209}

\bibitem{tsuang1990}
Tsuang, M.T., Lyons, M.J., Faraone, S.V.: {Heterogeneity of Schizophrenia:
  Conceptual Models and Analytic Strategies}. British Journal of Psychiatry
  \textbf{156}(1),  17--26 (1990). \doi{10.1192/bjp.156.1.17}

\bibitem{n4bias}
Tustison, N.J., Avants, B.B., Cook, P.A., Zheng, Y., Egan, A., Yushkevich,
  P.A., Gee, J.C.: N4itk: Improved n3 bias correction. IEEE Transactions on
  Medical Imaging  \textbf{29}(6),  1310--1320 (2010).
  \doi{10.1109/TMI.2010.2046908}

\bibitem{VanDenOord2017}
{Van Den Oord}, A., Vinyals, O., Kavukcuoglu, K.: {Neural discrete
  representation learning}. Advances in Neural Information Processing Systems
  \textbf{2017-Decem}(Nips),  6307--6316 (2017)

\bibitem{van2013wu}
Van~Essen, D.C., Smith, S.M., Barch, D.M., Behrens, T.E., Yacoub, E., Ugurbil,
  K., Consortium, W.M.H., et~al.: The wu-minn human connectome project: an
  overview. Neuroimage  \textbf{80},  62--79 (2013)

\bibitem{vaswani2017attention}
Vaswani, A., Shazeer, N., Parmar, N., Uszkoreit, J., Jones, L., Gomez, A.N.,
  Kaiser, {\L}., Polosukhin, I.: Attention is all you need. Advances in neural
  information processing systems  \textbf{30} (2017)

\bibitem{Vieira2019}
Vieira, S., Gong, Q.y., Pinaya, W.H.L., Scarpazza, C., Tognin, S.,
  Crespo-Facorro, B., Tordesillas-Gutierrez, D., Ortiz-Garc{\'{i}}a, V.,
  Setien-Suero, E., Scheepers, F.E., {Van Haren}, N.E.M., Marques, T.R.,
  Murray, R.M., David, A., Dazzan, P., McGuire, P., Mechelli, A.: {Using
  Machine Learning and Structural Neuroimaging to Detect First Episode
  Psychosis: Reconsidering the Evidence}. Schizophrenia Bulletin
  \textbf{46}(1),  17--26 (2019). \doi{10.1093/schbul/sby189},
  \url{https://doi.org/10.1093/schbul/sby189}

\bibitem{Walton2017}
Walton, E., Hibar, D.P., van Erp, T.G.M., Potkin, S.G., Roiz-Santia{\~{n}}ez,
  R., Crespo-Facorro, B., Suarez-Pinilla, P., {Van Haren}, N.E.M., de~Zwarte,
  S.M.C., Kahn, R.S., Cahn, W., Doan, N.T., J{\o}rgensen, K.N., Gurholt, T.P.,
  Agartz, I., Andreassen, O.A., Westlye, L.T., Melle, I., Berg, A.O.,
  M{\o}rch-Johnsen, L., Faerden, A., Flyckt, L., Fatouros-Bergman, H.,
  J{\"{o}}nsson, E.G., Hashimoto, R., Yamamori, H., Fukunaga, M., Preda, A.,
  {De Rossi}, P., Piras, F., Banaj, N., Ciullo, V., Spalletta, G., Gur, R.E.,
  Gur, R.C., Wolf, D.H., Satterthwaite, T.D., Beard, L.M., Sommer, I.E., Koops,
  S., Gruber, O., Richter, A., Kr{\"{a}}mer, B., Kelly, S., Donohoe, G.,
  McDonald, C., Cannon, D.M., Corvin, A., Gill, M., {Di Giorgio}, A.,
  Bertolino, A., Lawrie, S., Nickson, T., Whalley, H.C., Neilson, E., Calhoun,
  V.D., Thompson, P.M., Turner, J.A., Ehrlich, S.: {Positive symptoms associate
  with cortical thinning in the superior temporal gyrus via the ENIGMA
  Schizophrenia consortium.} Acta psychiatrica Scandinavica  \textbf{135}(5),
  439--447 (may 2017). \doi{10.1111/acps.12718}

\bibitem{welch2013dsm}
Welch, S., Klassen, C., Borisova, O., Clothier, H.: The dsm-5 controversies:
  How should psychologists respond? Canadian Psychology/psychologie canadienne
  \textbf{54}(3), ~166 (2013)

\bibitem{Wolfers2018}
Wolfers, T., Doan, N.T., Kaufmann, T., Aln{\ae}s, D., Moberget, T., Agartz, I.,
  Buitelaar, J.K., Ueland, T., Melle, I., Franke, B., Andreassen, O.A.,
  Beckmann, C.F., Westlye, L.T., Marquand, A.F.: {Mapping the Heterogeneous
  Phenotype of Schizophrenia and Bipolar Disorder Using Normative Models.} JAMA
  psychiatry  \textbf{75}(11),  1146--1155 (nov 2018).
  \doi{10.1001/jamapsychiatry.2018.2467}

\bibitem{ZHOU2005}
Zhou, S.Y., Suzuki, M., Hagino, H., Takahashi, T., Kawasaki, Y., Matsui, M.,
  Seto, H., Kurachi, M.: Volumetric analysis of sulci/gyri-defined in vivo
  frontal lobe regions in schizophrenia: Precentral gyrus, cingulate gyrus, and
  prefrontal region. Psychiatry Research: Neuroimaging  \textbf{139}(2),
  127--139 (2005). \doi{https://doi.org/10.1016/j.pscychresns.2005.05.005},
  \url{https://www.sciencedirect.com/science/article/pii/S0925492705000697}

\end{thebibliography}

\section{Supplementary Material}

\subsection{Datasets}
In this study, we used T1-weighted volumes from healthy subjects to train our normative models of the brain. These volumes were from two datasets: the Human Connectome Project - Young Adult (HCP-YA) \cite{van2013wu} and the Human Connectome Project - Development (HCP-D) \cite{somerville2018lifespan}. From the HCP-YA dataset, we used 1,113 volumes taken from the “1200 Subjects Data Release”\footnote[1]{http://www.humanconnectome.org/documentation/S1200/}. From the HPC-D, we used 652 volumes from the "Lifespan 2.0 Release" \footnote[2]{https://www.humanconnectome.org/study/hcp-lifespan-development/document/hcp-development-20-release}. In total, we had 1,765 subjects (808 male and 957 female) with an age range from 5 to 37 years old (Avg. (SD) =  23.3(8.1) years old). 

To evaluate our method, we used the Human Connectome Project - Early Psychosis (HCP-EP) ("Release 1.1"), a study with the goal to acquire high quality imaging, behavioral, clinical, cognitive, and genetic data on a cohort of early psychosis patients. Importantly, the data from HCP-EP was acquired at different acquisition sites. The HCP-EP focus on early psychosis (both affective and non-affective psychosis), within the first 3 years of the onset of psychotic symptoms. This is a critical time period when there are fewer confounds such as prolonged medication exposure and chronicity, and when early intervention strategies will be most effective. From the HCP-EP, we used the volumes of 46 healthy and 47 volumes of subjects with diagnosed with schizophrenia and balanced the groups for age and sex (see Table \ref{tab:demographic} for demographic details). 

\begin{table}[]
\caption{Demographic information for the subjects from the Human Connectome Project Young Adults (HCP-YA), Human Connectome Project Development (HCP-D), and Human Connectome Project for Early Psychosis (HCP-EP). For the HCPEP, we are presenting the data from the following classes: Control (HP) and subjects with early psychosis (EP). We used Student's t test and Chi-square test to verify if age and gender are significantly different in the HCP-EP dataset.}\label{tab:demographic}
\begin{center}
\begin{tabular}{l c c c c c}
\toprule
                    & \multirow{3}*{\shortstack{HCP-D\\(n=652)}}    & \multirow{3}*{\shortstack{HCP-YA\\(n=1,113)}}    & \multicolumn{3}{c}{HCP-EP (n=93)} \\ 
                    \cmidrule(lr){4-6}
                    &                                   &     & \shortstack{HP\\(n=46)}     & \shortstack{EP\\(n=47)} & stats \\ 
\midrule\midrule[.1em]
Age, y              & & & & & $1.87_{p=0.07}$ \\ 
\hspace{3mm}$Mean_{\pm SD}$         & $14.0_{\pm 4.1}$        & $28.8_{\pm 3.7}$       & $23.6_{\pm 2.8}$  & $22.6_{\pm 2.6}$  &   \\ 
\hspace{3mm}$Range$               & 5-21                & 22-37               & 16-30          & 19-31 &\\ 
Sex, n          & & & & & $2.07_{p=0.15}$\\ 
\hspace{3mm}Men$_{(\%)}$                 & $301_{(46\%)}$           & $507_{(45\%)}$          & $29_{(63\%)}$     & $37_{(79\%)}$  & \\ 
\hspace{3mm}Women$_{(\%)}$               & $351_{(54\%)}$           & $606_{(55\%)}$          & $17_{(37\%)}$     & $10_{(21\%)}$  & \\
\bottomrule
\end{tabular}
\end{center}
\end{table}

\subsection{MRI processing}
All images were corrected for intensity non-uniformity originating from the bias field using the function N4 bias-field correction \cite{n4bias} from the  Advanced Normalisations Tools (ANTs - version 2.3.4) \cite{ants}. The images were also  
registered to a common space (MNI152NLin2009aSym) applying rigid and affine transformations using the RegistrationSynQuick command from the ANTs. At the end, we had high-resolution volumes ($1mm^3$), where each volume had 192 x 224 x 192 voxels.

\subsection{Model Implementation}

\paragraph{VQVAE}
The VQVAE learns to map the 3-dimensional structural MRI data input into a discrete latent representation. It is constituted by an Encoder and a Decoder. The Encoder maps observations $x \in {\mathbb{R}}^{D}$ to a latent embedding $z \in {\mathbb{R}}^{d \times n}$, where $D$ and $d$ represent respectively the dimensions of the observations and the latent representations and $n$ represents the dimensions of the latent embedding vector. Each element in the latent embedding $z_e$ is mapped onto the nearest vector in a finite vocabulary codebook $e_K \in {\mathbb{R}}^{n}$ where $K$ is the pre-determined size of the discrete space. This discreteness allows the autoregressive Transformer to estimate the probability of any of the $K$ codebook vectors to be generated next. The Decoder reconstructs the original observation $\hat{x}^D$ from the latent embedding.

\paragraph{Autoregressive Transformer}
Autoregressive models explicitly learn the likelihood function of the next index by mapping the previous values in a sequence to its estimation of the probability of the possible values, $p(s_i) = p(s_i | s_{<i})$. We use a decoder-only transformer architecture due to its autoregressive nature and because it outperforms other autoregressive models such as the PixelCNN \cite{Pinaya2022}. The autoregressive transformer receives as input 1D sequences which are unmasked sequentially so that the model is only informed by values it has already estimated. The 3D latent representation is flattened using the raster scan order before being given as input to the transformer. The categorical nature of the VQ-VAE latent representation allows the transformer to predict the likelihood of any of the available elements in the codebook. This is done through a softmax non-linear function as the transformer output. The transformer is trained by maximizing the training log-likelihood, $ {\mathbb E}_{x \sim p(x)} \, [ -log \,p(s_i) ]$, of the training data, where only healthy participants are considered.

\paragraph{Implementation details}
Regarding the VQVAE architecture, the model was composed of 3 residual blocks, outputting 256 channels each. The discrete latent space was composed of 128 code elements where each was composed of a vector of length 64. The model was trained for 200 epochs with a batch size of 16, with the Adam optimiser, a learning rate of $3e^{-4}$ and a learning rate decay of 0.9999. Dropout at a rate of 0.1 was used.

Regarding the Transformer architecture, the autoregressive model was composed of 16 layers, with 8 attention heads. It receives as input the bottleneck from the VQVAE which is 16128 elements long. The vocabulary size is 128. The model was trained for 356 epochs with a batch size of 2. The Adam optimiser was used alongside a learning rate of $1e^{-3}$ and a learning decay of 0.999997. Dropout was used at a rate of 0.3.

\subsection{Baseline normative models}
\paragraph{Voxel-wise Brain age prediction with deep neural networks}
Brain age prediction consists in building predictive models of participants' age using only brain data, generally T1-weighted MRI scans. It uses multivariate regression to find correlates of changes in brain-structure and the biological age to build age predictors, despite the ageing process not being uniform across a population. As symptoms of psychiatric disorders are exacerbated during ageing, the brain age predictor trained on healthy control participants has been suggested as a normative model for psychiatric conditions, by measuring prediction error \cite{Cole2019, Marquand2019}. We used  the state-of-the-art in brain age prediction, the Simple Fully Convolution Network (SFCN) \cite{Peng2021}, trained on the HCP-D and HCP-YA 3-dimensional brain data, as a normative model of the brain trained on predicting age. The model's prediction error on the HCP-EP data was used as a proxy measure for detection of participants with schizophrenia.

\paragraph{Region-wise Gaussian Process Regression}
Following previous literature on phenotyping schizophrenia using normative models \cite{Wolfers2018}, we used set of Gaussian process regressors (GPR) trained to predict regional volumes of 183 brain regions using age and sex as covariables. In inference time, the trained models estimate the predicted brain volume and the confidence of prediction. The z-scores (i.e., the prediction error normalized by the prediction uncertainty) across all regions for each participant are used as proxies of anomaly. In this study, we used the Predictive Clinical Neuroscience (PCN) toolkit \footnote[4]{github.com/amarquand/PCNtoolkit} that is optimised for normative modelling of clinical imaging data.

\paragraph{Region-wise Bayesian Linear Regression}
As a third baseline, we used Bayesian linear regressive (BLR) models, trained to infer the same regional volumes from healthy control participants using age and sex as covariables \cite{HUERTAS2017134}. Similar to our approach and unlike Gaussian process regressors, Bayesian linear regressors estimate likelihood-based statistics that can be used as proxies for normative models. They are able to model non-Gaussian predictive distributions, as is the case in our approach. As in the Gaussian process regression, we measure as proxy of anomaly detection the mean z-score over all regions measured for a given participant using the PCN toolkit. The larger the z-score, the more out of distribution the sample is.

\subsection{Region-level analysis - extension}

\begin{figure}
\centering
\includegraphics[width=\textwidth]{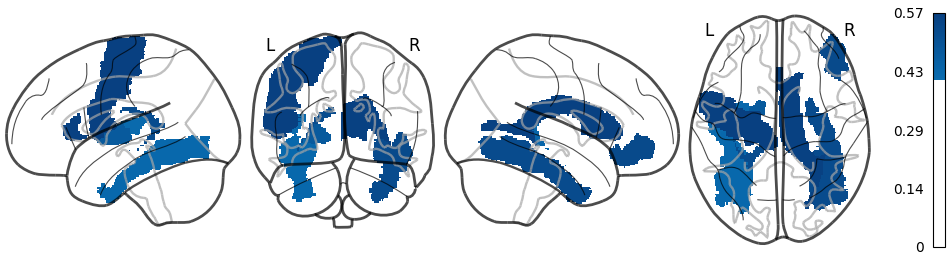}
\caption{Top 10 brain regions with the highest effect size measured by Cohen's d.}
\label{fig:fig2}
\end{figure}

We estimate the normative scores of each cortical region and anatomical structure (from aseg+aparc parcellation) by calculating the median log-likelihood of the latent variables that are inside the region when upsampling them onto the original brain space. We found that 16 regions out of 113 presented a different normative score between control and subjects with schizophrenia with a significance level below p=0.05, but none show significance once the result is corrected by the Bonferroni correction for multiple comparisons \cite{dunn1961multiple}. These results contrast with the global estimation where the difference between log-likelihoods of the two cohorts is significant (p-value<0.001). The Bonferroni correction is quite conservative and assumes independence between estimations, which in statistical inferences in homotopic and adjacent regions is not a correct assumption. Instead, here we focus on the effect sizes of each statistical test. In Fig. \ref{fig:fig2}, we show the regions with highest measure effect size (Cohen's d), with them presenting small to medium effect sizes (between 0.42 and 0.57). The regions with the highest values were present in the prefrontal cortex (i.e., the left precentral gyrus and the right pars orbitalis), the temporal cortex (i.e., the right and left fusiform gyri and the left transverse temporal gyrus), the right lateral ventricle, the anterior portion of the corpus callosum, and the left and right choroid plexus.

\end{document}